\documentclass[runningheads]{llncs}
\usepackage{graphicx}
\usepackage{subfigure}
\usepackage{siunitx}
\usepackage{amsmath}

\numberwithin{equation}{section}

\begin{document}
\title{A GLCM Embedded CNN Strategy for Computer-aided Diagnosis in Intracerebral Hemorrhage}

\author{Yifan Hu \and
	Yefeng Zheng}
\authorrunning{F. Author et al.}
\institute{Tencent, Shenzhen 518057, CHINA \\
	\email{ivanyfhu@tencent.com}\\
	}

\maketitle              % typeset the header of the contribution
\begin{abstract}
Computer-aided diagnosis (CADx) systems have been shown to assist radiologists by providing classifications of all kinds of medical images like Computed tomography (CT) and Magnetic resonance (MR). Currently, convolutional neural networks play an important role in CADx. However, since CNN model should have a square-like input, it is usually difficult to directly apply the CNN algorithms on the irregular segmentation region of interests (ROIs) where the radiologists are interested in. In this paper, we propose a new approach to construct the model by extracting and converting the information of the irregular region into a fixed-size Gray-Level Co-Occurrence Matrix (GLCM) and then utilize the GLCM as one input of our CNN model. In this way, as an useful implementary to the original CNN, a couple of GLCM-based features are also extracted by CNN. Meanwhile, the network will pay more attention to the important lesion area and achieve a higher accuracy in classification. Experiments are performed on three classification databases: Hemorrhage, BraTS18 and Cervix to validate the universality of our innovative model. In conclusion, the proposed framework outperforms the corresponding state-of-art algorithms on each database with both test losses and classification accuracy as the evaluation criteria.

\end{abstract}
\section{Introduction}
Intracerebral hemorrhage is a worldwide disease with high occurence and mortality. An early and accurate diagnosis is of highly importance for the patients otherwise a severe consequence will come down. As a quick tool to detect the location and quantity of the hemorrhage, rapid CT scan is used to determine cause of the hemorrhage thereafter a proper treatment is made. Thus, an accurate medical image classification algorithm on rapid CT scan will be valuable for both patients and doctors as the reference. 

In recent years, a tremendous interest of CNNs has arisen in the field of computer vision and medical image processing. For example, it is shown that CNN models can achieve a comparable performance to experienced radiologists in lung cancer predication [1] and retinal disease diagnosis [2]. For hemorrhage tasks, an ROI, either a segmentation or a detection box of the hemorrhage, is usually provided by radiologists where we should pay more attension to. Clinically, a diagnosis will be made according to the texture and morphology inside the ROI. However, it is difficult for a CNN model to serve so many shapeless hemorrhage (ROI), especially for the small hemorrhage region, as the training set. Relying on the segmentation information of radiologists, designing an ROI-reachable CNN algorithm will be more reasonable for the classification task in estimation of the pathological cause of hemorrhage.

Although CNN has achieved so many breakthroughs, radiomics and handcrafted feature analysis are still playing another important role in CADx. Distinguished from CNN with the whole image as input, features and analysis in radiomics are only extracted based on ROI/segmentation area. As descriptors of the relationships between image voxels, gray-level co-occurrence matrix (GLCM) based features are one group of widely used handcrafted features in radiomics and CADx [3-4]. Relationships of image voxels within the ROI are calculated in GLCM features while only intensity based features can be computed from CNN models, that is to say, GLCM based features could be a perfect complementary to CNN features. Some papers [5-6] claimed that they achieved a better performance in classification by utilizing a combination of deep and handcrafted features. However, it is still essential to manually choose some proper features after the combination for these methods [5-6]. On the contrary, sometimes it will lead to a decrease on the classification because a lot of redundant information are produced after the fusion of CNN and handcrafted features.  

\begin{figure}
	\centering
	\includegraphics[width=10cm]{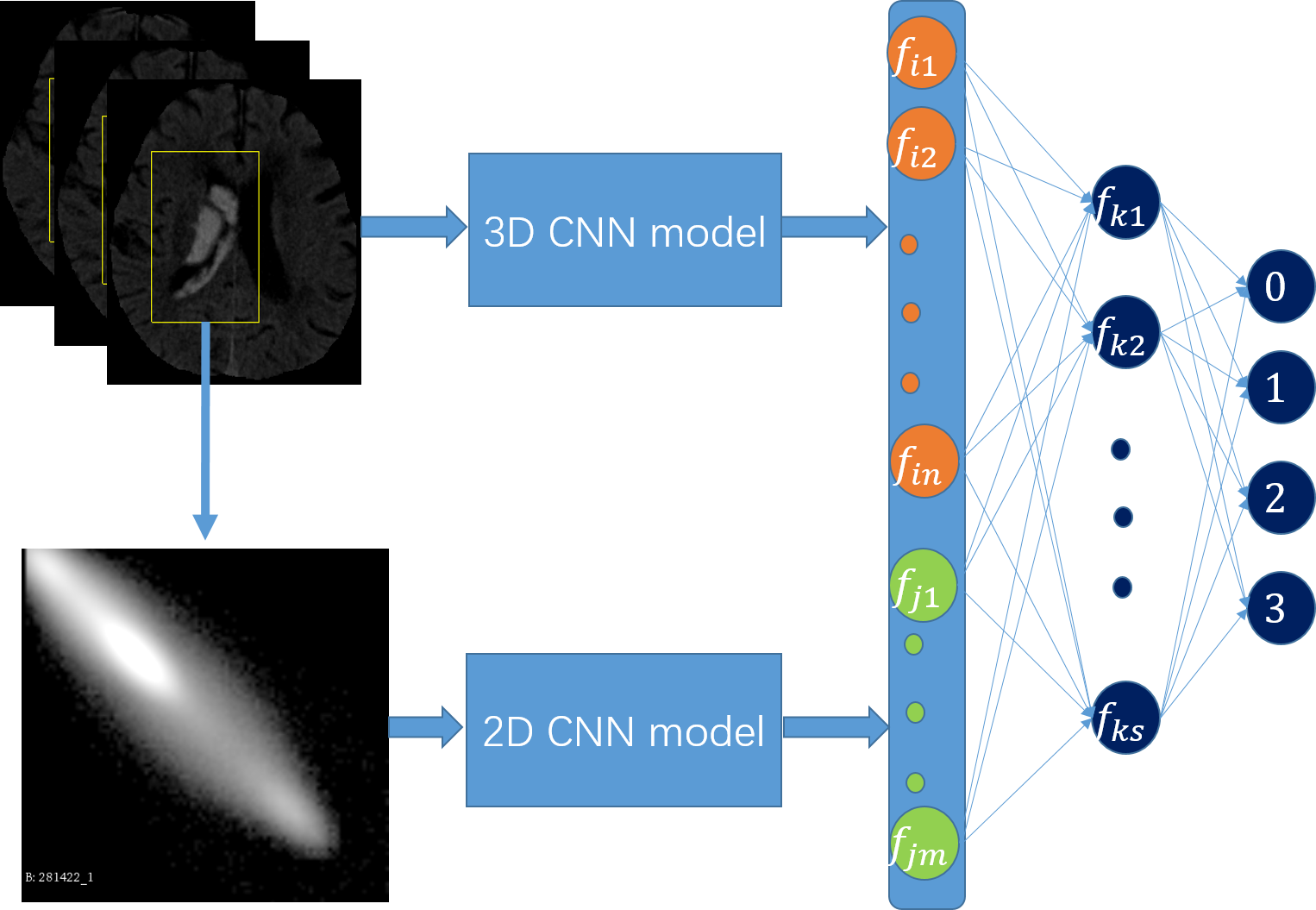}
	\caption{Architecture of our proposed GLCM-CNN, a hybrid model with two inputs: whole image and GLCM image from irregular region.} \label{fig1}
\end{figure}

In this paper, we propose a strategy that can implant the GLCM into any CNN model, called GLCM-CNN, where CNN can take advantage of the information in irregular ROIs. Meanwhile, no feature selection is needed for our model since the CNN can automatically decide the optimal feature set. The main innovations of this paper are listed as follows:

1. Convert the image within ROI into the "GLCM image" as a domain transformation.

2. Consider the  ``GLCM image" as part of the input of the CNN model and then make the original image and ROI information be trained together and therefore achieve a better result in classification of pathological cause of hemorrhage.

3. Test this hybrid model on two backbone CNN models and two more publicly available databases to prove the universality of this strategy.

\section{Methodology}
The proposed strategy consists of two branches (see Figure~\ref{fig1}). The branch on the top, which could be any widely used CNNs for classification tasks, is responsible for extracting the global features from the whole image. In this section, we will detailedly introduce the ``GLCM" branch, including the construction of the GLCM image and the integrating of these two branches as the GLCM-CNN.
\subsection{Gray-level co-occurrence matrix (GLCM)}
GLCM computes all frequencies of intensity pairs occurred in an image and then each frequency value is recorded in its corresponding element. For example, the frequency of pair (1,3) will become the element of row 1 and column 3 in GLCM. Furthermore, along with 4 directions (\ang{0}, \ang{45}, \ang{90}, \ang{135}), 4 different GLCMs could be derived from a 2D image. Traditionally, a couple of manual statistics will be computed from GLCMs as the texture features of image. However, it is impossible to design a feature that works well on all classification tasks. Thus, usually a feature selection step is also necessary to choose a set of proper features for a specific task.

\begin{figure}[htbp]
	\centering
	\subfigure[]{
	\includegraphics[width=5.5cm]{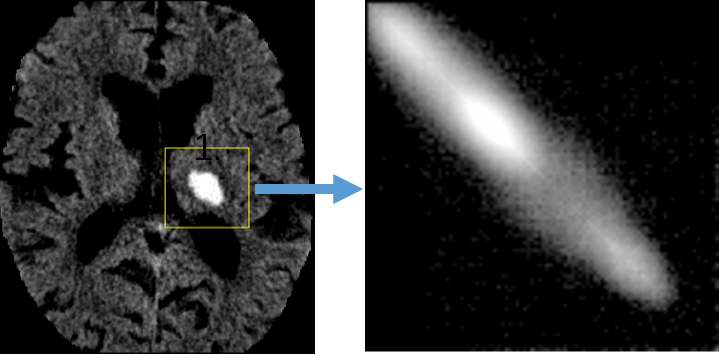}
    }
    \quad
    \subfigure[]{
    \includegraphics[width=5.5cm]{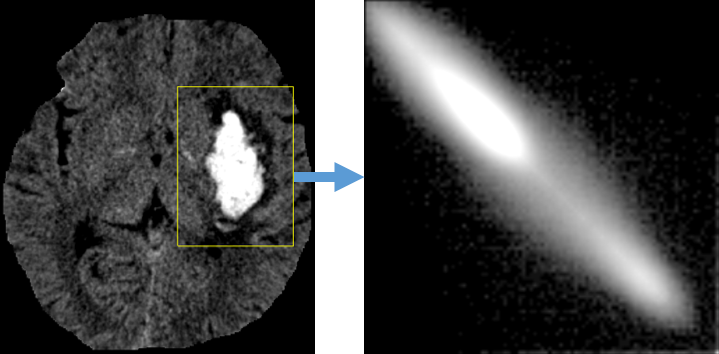}
    }
    \quad
    \subfigure[]{
    	\includegraphics[width=5.5cm]{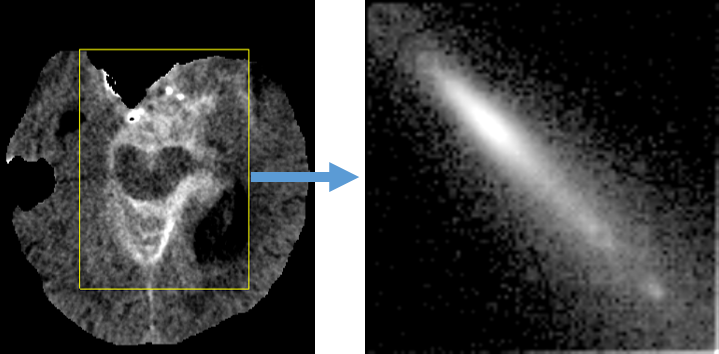}
    }
	\caption{(a) and (b) show that the similarity of inter-class data (hemorrage caused by hypertension) in spite of the size variation of the ROIs; (b) and (c) show that the difference between intra-class data (hemorrage caused by hypertension and aneurysm, respecitively).} \label{fig2}
\end{figure}
 
\subsection{Construction of GLCM Image}
As an important property of GLCM, this transformation can be processed with any input image no matter what shape/size of the image has. [3] From the definition it is shown that the output size only depends on the intensity level of the input image. Moreover, GLCM will pay more attension on the texture of the input region instead of simple intensity statistics (e.g. histogram). (see Figure~\ref{fig2})  Thus, in order to keep the same GLCM sizes, a normalization could be taken for a dataset and usually an intensity range of [0, 255] is restricted after the normalization. But there are still multiple cases that you need deal with the normalized image differently depending on different types of medical image:

(1) 2D image: 4 fixed size GLCMs correspongding to the ROI will be obtained after a normalization and an average of all GLCMs is calculated as the GLCM image. 

(2) Isotropic 3D image: 13 fixed size GLCMs correspongding to the ROI will be obtained after a normalization and an average of all GLCMs is calculated as the GLCM image.

(3) 3D images with different resolutions in z-axis: calculate the GLCM image slice by slice in the x-y plane and add them together.

(4) Images with multi-channels (MRIs or RGB medical images): calculate the GLCM image for each channel and combine them together as a multi-channel GLCM image.

For the 3 experiments in Section 3, the derived GLCM image size of hemorrage dataset
is $96 \times 96$ because of the fewer intensity levels of the CT image; the GLCM image size of MRI dataset is $256\times256\times4$ and the GLCM image size of RGB dataset is $256\times256\times3$.

\subsection{Training a CNN with GLCM Image}
After the GLCMs are generated, traditionally, features are designed based on all element values. For example, contrast feature and homogeneity feature are listed as follows:

\begin{subequations}
	\begin{align}
	Contrast=\sum_{n=1}^{N-1}P_{ij}(i-j)^2\\     Homogeneity=\sum_{n=1}^{N-1}\frac{P_{ij}}{1+(i-j)^2}
	\end{align}
\end{subequations}
As can be seen, lots of features can be designed as a linear combination of all elements of the GLCM. However, no theoretical proof is provided that an optimal classification result can be achieved via the proposed features. Therefore, rather than extracting features manually, it is better to consider the GLCM as an image and use CNN to derive the optimal features. Compared with traditional features, there are several advantages for CNN features: 1) The coefficient of traditional features may not be the optimal while CNN could obtain optimal parameters by optimizing the loss; 2) Only linear operator is occurred in constructing traditional features while CNN has some non-linear oprations like pooling and activation; 3) A feature selection is required for feature engineer but CNN can provide an optimal set of features simultaneously.

As illustrated in the flowchart of Figure~\ref{fig1}, the whole image and the GLCM image are trained together to obtain the model. Similar to all CNN models, the sizes and parameters of the GLCM-CNN can be adjusted for training. In GLCM-CNN, the features extracted from GLCM image, which focus on the intensity correlation only in ROI and then reflect more texture details, will be a superb supplementary to the CNN features. For example, it might be a hard task for regular CNN to pay attention on each ROI region if a database has both large and small lesions. Thus, with a forcible feature extraction with GLCM image on ROI, the information of ROI is amplified and help the model improve the classification performance.

In this paper, we regard two widely used classification models, ResNet-18 [7] and VGG-11 [8], as our backbone models. To be compared, the GLCM image branch is added to the backbone models, respectively, and the result is illustrated in Section 3. We perform the experiment in PyTorch and Titan P40 GPUs. The adam algorithm and cross entropy loss is adopted to optimize the network and learning rate is initialized to be 0.05. The network weights are initialized by Kaiming He's algorithm. In total of 50 epochs are trained for each algorithm.

\section{Experiments and Results}
\subsection{Database for the Experiment}
\subsubsection{Intracerebral hemorrhage} Released and well labelled by radiologists and physicians, this database include in total 1476 intracerebral hemorrhage CT images, which are subdivided into 4 pathological causes of hemorrhage: 784 aneurysm (An) data, 570  hypertension (Ht) data, 97 arteriovenous malformation (AVM) data and 36 Moyamoya disease (MMD) data.
\subsubsection{BraTS18} This is an open sourced database including 163 MRI data with a prediction survival label for patients with brain tumor: short, medium and long. [9-10] 4 modals of MRI are provided by the organizers, which are Flair, T1, T1ce and T2.
\subsubsection{Cervix} This is another publicly available database published in Kaggle competition. [11] This database aim to classify cervix types based on cervical images. 1431 cervical RGB images with 288 Type\_1 images, 689 Type\_2 images and 454 Type\_3 images are prepared for the experiments.

\subsection{Comparison with State-of-Art Algorithms}
\subsubsection{Implementation Details}
As we described above, we compare our algorithm with its corresponding backbone network to show the advantages of our proposed model. The CT images in our dataset are rapid CT scans taken from year 2015 to 2018. The CT scans were acquired with a slice distance of 4–7 mm and an in-plane resolution of 0.5-0.9. Since the similarity on the size of human head, all data are resized to a $230\times270\times30 mm{^3}$ volume. Two widely-used classification CNNs are served as our backbone network: ResNet-18 and VGG-11, and at last layer 1024 features are extracted by each of them, while our model has 32 features in addition. In this paper, we use another ResNet style CNN and the input GLCM image size is $96\times96$ for the glcm branch. In output layer, in total of 1024+32 features are processed by a softmax function and classified into 4 classes.
 
\subsubsection{Experiment Results on hemorrhage database}
A 5-fold cross validation is explored to evaluate our model. Each fold includes 20\% of all data, where the propotion of all classes are the same as the whole database. In order to evaluate the performance, we compute average cross entropy loss, accuracy and AUC for each class (one-vs-others) of the test set. Table~\ref{tab1} quantitatively demonstrates the optimal test loss with its corresponding accuracy and 4 AUCs of 4 classes.

\begin{table}
	\centering
	\caption{Comparison of the average classification performances with different methods on the hemorrhage dataset.}\label{tab1}
	\begin{tabular}{|l|l|l|l|l|l|l|}
		\hline
		Method &  Loss & Acc & AUC1 & AUC2 & AUC3 & AUC4\\
		\hline
		ResNet18 & 0.8839 & 0.8607 & 0.9631 & 0.9575 & 0.8122 & 0.7137\\
		ResNet18+GLCM &  \textbf{0.8697} & \textbf{0.8755} & \textbf{0.9654} & \textbf{0.9607} & \textbf{0.8353} & \textbf{0.7354}\\
		VGG11 & 0.9110 & 0.8371 & 0.9442 & 0.9413 & \textbf{0.7680} & 0.7443\\
		VGG11+GLCM & \textbf{0.8984} & \textbf{0.8506} & \textbf{0.9568} & \textbf{0.9444} & 0.7140 & \textbf{0.7544}\\
		\hline
	\end{tabular}
\end{table}

Compared to both of two backbone models, our GLCM-CNN models perform better on all evaluation scores. The average loss and accuracy of our model has achieved 0.8697 and 0.8755, which have gains of 0.0142 and 0.0148, respectively. Especially, AUCs of two small classes AVM and MMD, which only have 97 and 36 data, respectively, of 1486, have improved with more than 2\%. This observation will be beneficial for radiologists who might be less experienced on these two rare diseases.

\begin{figure}[htbp]
	\centering \vspace{-5mm}
	\subfigure[]{
		\includegraphics[width=5.5cm]{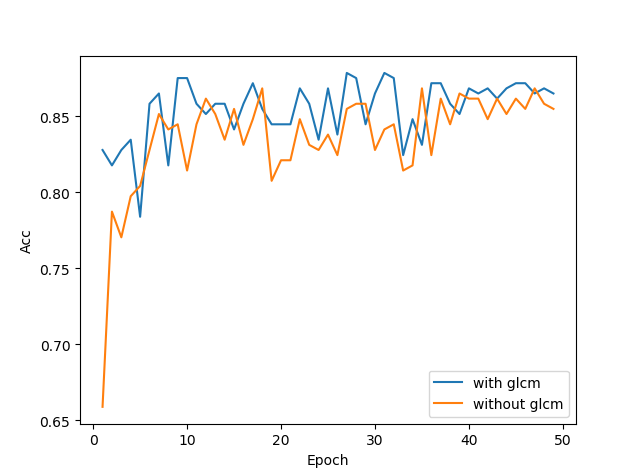}
	}
	\quad
	\subfigure[]{
		\includegraphics[width=5.5cm]{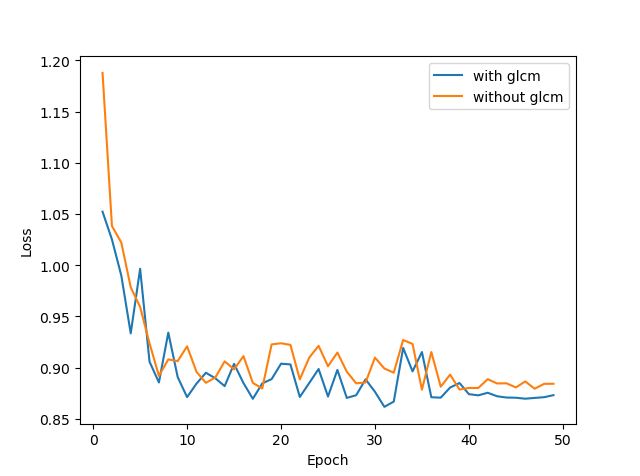}
	}
	\caption{(a) shows the accuracy of each epoch for one fold of the 5 cross validation experiment with hemorrage dataset for both GLCM-CNN and CNN; (b) shows the loss of each epoch of the experiment for both GLCM-CNN and CNN.} \label{fig3}
\end{figure}

\vspace{-5mm}
Figure~\ref{fig3} illustrate the training and test loss within 50 epochs for one fold of our experiment. As can be seen, compared with the backbone model, the blue line of our model consistently has a higher accuracy and lower loss in the experiment. Moreover, the accuracy and the loss of our model converge quickly and perform steadily, which indicates that our model can achieve a good accuracy in a short time. 

\subsubsection{Experiment Results on other database} 
In order to validate the effectiveness of our algorithm, two more databases are also explored to compare with the state-of-art CNN models. The sampling strategy are almost the same as the hemorrhage and the only difference is that 24 features are obtained for BraTS dataset instead of 32 features. The reason is to prevent overfitting because of the small database.

\begin{table}
	\centering
	\caption{Comparison of the average classification performances with different databases.}\label{tab2}
	\begin{tabular}{|l|l|l|l|l|}
		\hline
		Method &  Loss\_BraTS & Acc\_BraTS & Loss\_Cervix & Acc\_Cervix \\
		\hline
		ResNet18 & 1.0407 & 0.5118 & 0.9722 & 0.5756 \\
		ResNet18+GLCM & \textbf{1.0248} & \textbf{0.5264} & \textbf{0.9670} & \textbf{0.5802}\\
		VGG11 & 1.0209 & \textbf{0.5408} & 0.9495 & 0.6049\\
		VGG11+GLCM & \textbf{1.0183} & 0.5356 & \textbf{0.9485} & \textbf{0.6089} \\
		\hline
	\end{tabular}
\end{table}

It is shown in Table~\ref{tab2} that the performances of our proposed models are better than backbone models, individually, no matter for the loss or the accuray. Furthermore, the ResNet18, which is a more complicated model, benifits more from the GLCM branch than the simpler model VGG11 does. That is to say, our proposed model can reduce the overfitting for a complex model and keep the accuracy in the meantime.

\section{Discussion}
In this paper, we propose a novel CNN method to perform classification on medical images. With more concentration on ROI area, our innovative CNN model exploits the ROI information by converting the irregular ROI into a GLCM image, which is part of the input for the CNN model. We attempt two widely used CNN model as our backbone models and three databases with 3 different kinds of medical images to validate the effectiveness and universality of our strategy. Good performances are offered by our model, compared to two state-of-art models, in all of 3 databases which can be found in Table~\ref{tab1} and Table~\ref{tab2}.

Only the GLCM image is involved in our model in this paper, but there are also other types of handcrafted features like Grey-Level Run Length Matrix (GLRLM) and Gray level size zone matrix (GLSZM) which can be explored similarly in the future. A proper balance between the number of GLCM features and CNN features will also be a problem for future research.

\end{document}